\documentclass[review,12pt]{elsarticle}
\usepackage{amssymb}
\usepackage{amsfonts}
\usepackage{amssymb}
\usepackage{mdwmath}
\usepackage{mdwtab}
\usepackage{natbib}
\usepackage{color}
\usepackage{amsmath}
\usepackage{subfigure}
\usepackage{url}
\usepackage{graphicx}
\usepackage{listings}
\usepackage{soul}

\journal{Applied Soft Computing}
\begin{document}
\begin{frontmatter}

\title{Genetic Programming  for Smart Phone Personalisation}
\author{Philip Valencia, Aiden Haak, Alban Cotillon,   and Raja Jurdak}
\address{
 Autonomous Systems Laboratory\\
CSIRO Computational Informatics, Brisbane Australia\\
}
\begin{abstract} Personalisation in smart phones requires adaptability to dynamic context based on user mobility, application usage and sensor inputs. Current personalisation approaches, which rely on static logic that is developed a priori, do not provide sufficient adaptability to dynamic and unexpected context. This paper proposes genetic programming (GP), which can evolve program logic in realtime,  as an online learning method to deal with the highly dynamic context in smart phone personalisation. We introduce the concept of collaborative smart phone personalisation through the GP Island Model, in order to exploit shared context among co-located phone users and reduce convergence time. We implement these concepts on real smartphones to demonstrate the capability of personalisation through GP and to explore the benefits of the Island Model. Our empirical evaluations on two example applications confirm that the Island Model can reduce convergence time by up to two-thirds over standalone GP personalisation.    \end{abstract}
\end{frontmatter}
\section{Introduction}

Smartphones have experienced exponential growth in recent years. These phones embed a growing diversity of sensors, such as gyroscopes, accelerometers, Global Positioning Systems (GPS), and cameras, with broad applicability in areas such as urban sensing or environmental monitoring.  Coupled with phone users' high mobility and diverse profiles~\citep{falaki}, this  sensory richness has fuelled complex applications with composite logic, including features such as location-based and usage-based services. The increased application complexity  involves significant challenges in personalising smart phone applications so that they can adapt to new or unexpected context.

Smartphone personalisation either occurs centrally at a server or locally at the phone. Centralised approaches track user activity to customise content delivery or application behaviour. They can easily misrepresent user preferences due to spurious activity, and they involve privacy concerns as users need to share their data with the content providers. Most  standalone smartphone algorithms, aiming at either data-centric~\citep{miele} or user-centric personalisation \citep{dockhorn}, are based on static or rule-based approaches. However, personalisation increasingly depends on contextual information and user inputs \citep{bae}, which are both subject to dynamic changes arising from mobility and user preferences. The problem of personalisation of smart phones is therefore multidimensional and requires an approach that can not only adapt parameters in response to changes in user preferences and context, but also adapt the program logic to optimise for unpredictable changes in context. 

Online learning is well-suited for smart phone personalisation. In order to provide maximum versatility to support the creativity and unpredictability of smart phone users, we need an online learning approach that allows for the adaptation of program logic, and not just parameters within  fixed program logic. Additionally, smart phone users are often co-located with several other users that share their context, providing opportunities for collaborative personalisation based on this shared context. The most suitable online learning strategy should provide the syntactic richness to evolve logic on a single smartphone, and to support the sharing of functional logical blocks among co-located phones according to the building block hypothesis~\cite{Holland75}. 

Most online learning methods, such as reinforcement learning~\cite{sutton84} and neural networks~\cite{hopfield},  are concerned with parameter optimisation only. While these methods could be run within an evolutionary framework, any sharing of genetic sequences across phones would not necessarily map to functional logic blocks, which could slow down convergence. Genetic Programming~(GP), on the other hand, is amenable to this scenario, as it evolves both logic and parameters simultaneously, providing it with the syntactic richness and flexibility that is comparable to offline human software development. Because GP evolves functional logic blocks, sharing logic across multiple phones is both more meaningful and more conducive to quicker convergence. 

This paper proposes GP for smart phone personalisation. We first show empirically  that GP can support smart phone personalisation through our software framework called the Android Genetic Programming Framework (AGP). In order to expedite convergence towards high performing applications, we propose collaborative personalisation of smartphones through the GP Island Model to exploit shared context between co-located phones. The main research question we aim to answer is: ``To what extent does collaborative smartphone personalisation improve convergence time?''. To address this question, we extend AGP to support the island model and run extensive experiments with 2 Android phones for a comparative evaluation of the benefits of sharing logic among smartphones against \emph{stand-alone} personalisation, exploring 6 scenarios with different migration rates and intensities. Our results show that the Island Model consistently outperforms the standalone GP  for online personaliation, and somewhat surprisingly, that injecting random programs into the subpopulations can be beneficial for the more complex application of energy-efficient localisation. 


The contributions of this paper are:

\begin{itemize}
\item Introduction, motivation, and demonstration of genetic programming for smart phone personalisation
\item Proposal of collaborative smart phone personalisation through the GP Island Model for faster convergence and exploitation of shared context 
\item Empirical evaluation of both standalone and collaborative personalisation through two case study applications, which confirm the benefits of collaborative personalisation
\end{itemize}

 The remainder of the paper is organised as follows. Section~\ref{sec:related} discusses related work in the literature.   Section~\ref{sec:prob} defines the problem for smart phone personalisation, motivates genetic programming to address this problem, and proposes collaborative GP for faster personalisation. Section~\ref{sec:agp}  briefly introduces the AGP framework and our extension to support the Island Model. Section~\ref{sec:perf} demonstrates online personalisation through GP, while Section~\ref{sec:island_exp}  evaluates the benefits of collaborative personalisation through the Island Model. Section~\ref{sec:conc} discusses the results and concludes the paper.


\section{Related Work}
\label{sec:related}
Most online learning approaches, including neural networks, adaptive systems, and reinforcement learning, use specialised structures for representation within the learning process~\cite{koza}. Because of their reliance on specialised structures, these approaches are amenable for online learning situations where the overall program logic is well-defined while individual parameters within this program structure need to be optimised. While these approaches have been used in an evolutionary context (for e.g. in~\cite{Xin93}), their representation of genetic material is not necessarily aligned to functional logic blocks, which makes them less amenable to logic sharing among islands. In contrast, genetic programming uses program logic representation in the learning process, supporting high versatility for entirely new actions in response to unpredictable stimuli. Through its program tree representation, GP can isolate functional logic blocks for sharing across islands to leverage shared context for faster convergence. 

Several previous GP works in  have  focused on architectural issues. Whereas some solutions provide generic frameworks for evolutionary computation problems \citep{mcphee,gagne,ventura}, others propose application-specific solutions. Ismail et al. \citep{ismail} describe a GP framework for extracting a mathematic formula needed for fingerprint matching, whereas, the authors in  \citep{torres} focus on a genetic programming framework for content-based image retrieval. In  \citep{lacerda}, Lacerda et al. introduce a framework for associating ads with web pages based on GP. Valencia et al. \citep{valencia} study genetic programming for Wireless Sensor Networks and propose the In Situ Distributed Genetic Programming (IDGP) framework. DGPF \citep{weise} brings utilities for Master/Slave, Peer-to-Peer, and P2P/MS hybrid distributed search execution. P-Cage \citep{folino} introduces and evaluates a complete framework for the execution of genetic programs in a P2P environment. It shows the relevance of using P2P networks scalability to counteract computation limitations. 

Design patterns describe the interaction between groups of classes or objects. They concentrate on specific concerns for implementing source code to support program organization. When they are well integrated into a framework, they ensure the goals of extensibility and reuse.  Lenaerts and Manderick \citep{lenaerts} discuss the construction of an object-oriented Genetic Programming framework using design patterns to increase flexibility and reusability. McPhee et al. \citep{mcphee} extend the latter to Evolutionary Computation (EC). As the  solution search space  for a problem becomes wider, it leads to a more abstract set of classes. Based on those works, Ventura et al. introduced JCLEC \citep{ventura}, a Java Framework for evolutionary computation. They present a layered architecture and provide a GUI for EC.  This paper similarly uses Java for a genetic programming framework,  albeit for a more resource constrained smart phone platform.

Smartphone personalisation research has mainly focused on rule-based approaches.  Korpipaa et al.~\citep{korpipaa} introduced a first framework for user customization using context changes as triggers. Their prototype enables the end-user to set up actions on context events such as GUI interactions, RFID tag informations, accelerometer peaks and other data from embedded sensors.
Since this first try on Nokia N73 smartphone, similar applications have been published on both Android and Apple applications stores. On{x} \citep{onx} and Launch Center Pro \citep{launchCenter} let users build rules for automating various tasks. These mechanisms still require explicit user involvement in personalisation, which limits their utility to more technology savvy users. 

More recently, Interactive Differential Evolution (IDE) has been applied on smart phones \citep{lee} as a method to achieve quick image enhancement of photos taken on mobile phones. As with most Interactive Evolutionary Computation (IEC) implementations, IDE encodes parameters of a fixed-logic solution as a vector and optimises these parameters over time, unlike GP which also optimises the logic. 

IEC methods include interactive evolution strategy~\citep{herdy}, interactive genetic algorithm~\cite{caldwell}, interactive genetic programming~\cite{sims}, and human-based genetic algorithm~\cite{kosorukoff}.
An interactive genetic algorithm (IGA) is defined as a genetic algorithm that uses human evaluation.

The community-based earthquake detection technique, proposed by Faulkner et al \cite{Faulkner2011}, highlighted the need to consider distributed context, using  accelerometer data from multiple smartphones to detect earthquakes. The distributed nature of the data  allowed the detection and filtering out of false positives and negatives due to spurious sensor data. This highlights the importance of distributed context for the  collaborative learning method in this paper.

Another instance of related work is activity recognition using accelerometers data \cite{Weiss2012}. Interestingly, Weiss found that using recognition models tailored specifically to a user outperformed an impersonal model which used data gathered from multiple users. The personal model also outperformed a hybrid model using a combination of a model based on data from both a specific user and a model based on data gathered from multiple users. However, it must be noted that these models were all developed offline and did not adapt to their users.  The work in this paper postulates and shows that collaborative learning can be useful for adapting to shared context for some applications, while other problems such as motion models may indeed be more amenable to personalised models, where standalone GP could learn the most appropriate logic. 

In an earlier paper, we introduced the Android Genetic Programming Framework~(AGP)~\cite{cotillon} as the first  genetic programming solution available on smartphones, demonstrating its ability to provide and update context-specific solutions over time. This paper proposes collaborative smart phone personalisation through the GP Island Model and extends AGP to evaluate this concept~\footnote{AGP source code and documentation are publicly available for download at http://sourceforge.net/projects/agpframework/}. 

\section{Smart Phone Personalisation through GP}
\label{sec:prob}
This section first defines the smart phone personalisation problem, and then motivates online learning, particularly through GP, to address this problem. The final part of the section argues for using collaborative learning through the Island Model to exploit shared context and shorten convergence time. 

\subsection{Problem Definition}
Mobile phone users  have always tried to customise their devices, for instance through personalised ring tones. The emergence of smartphones takes the personalisation possibilities to a new level. First of all, smartphones have access to a broad range of Internet data which can be augmented with sensor-based context information. Secondly, the  higher computing performance of smart phones enables developers  to create novel applications.
The combination of processing power,  content accessibility and context awareness opens new opportunities for  personalisation. Finally, the intrinsic smart phone user mobility dictates a highly dynamic and unpredictable context.  

Personalisation mechanisms have been slow to respond to these opportunities, with applications relying largely on static logic designed by the application programmer. These approaches are unable to cope with unexpected changes in context that is not specified in their static logic, highlighting the need for more dynamic personalisation methods.

The main problem that we are interested in this paper is to support continuous personalisation of smart phone applications in response to changes in environmental context and in user preference. Given that these changes are unpredictable at the time of application development, we propose online learning to address this problem. 

\subsection{Online Learning for Personalisation}
Several machine learning approaches, such as reinforcement learning, neural networks, support vector machines, and genetic programming,  can support online learning through integration into an evolutionary framework that iterates through their optimisation process. We compare the suitability of 2 candidate online learning approaches for smart phone personalisation in further detail: (1) neural networks; and (2) genetic programming.

\begin{figure}
\subfigure[Program tree representation]
{\centering
\includegraphics[width=0.5\columnwidth]{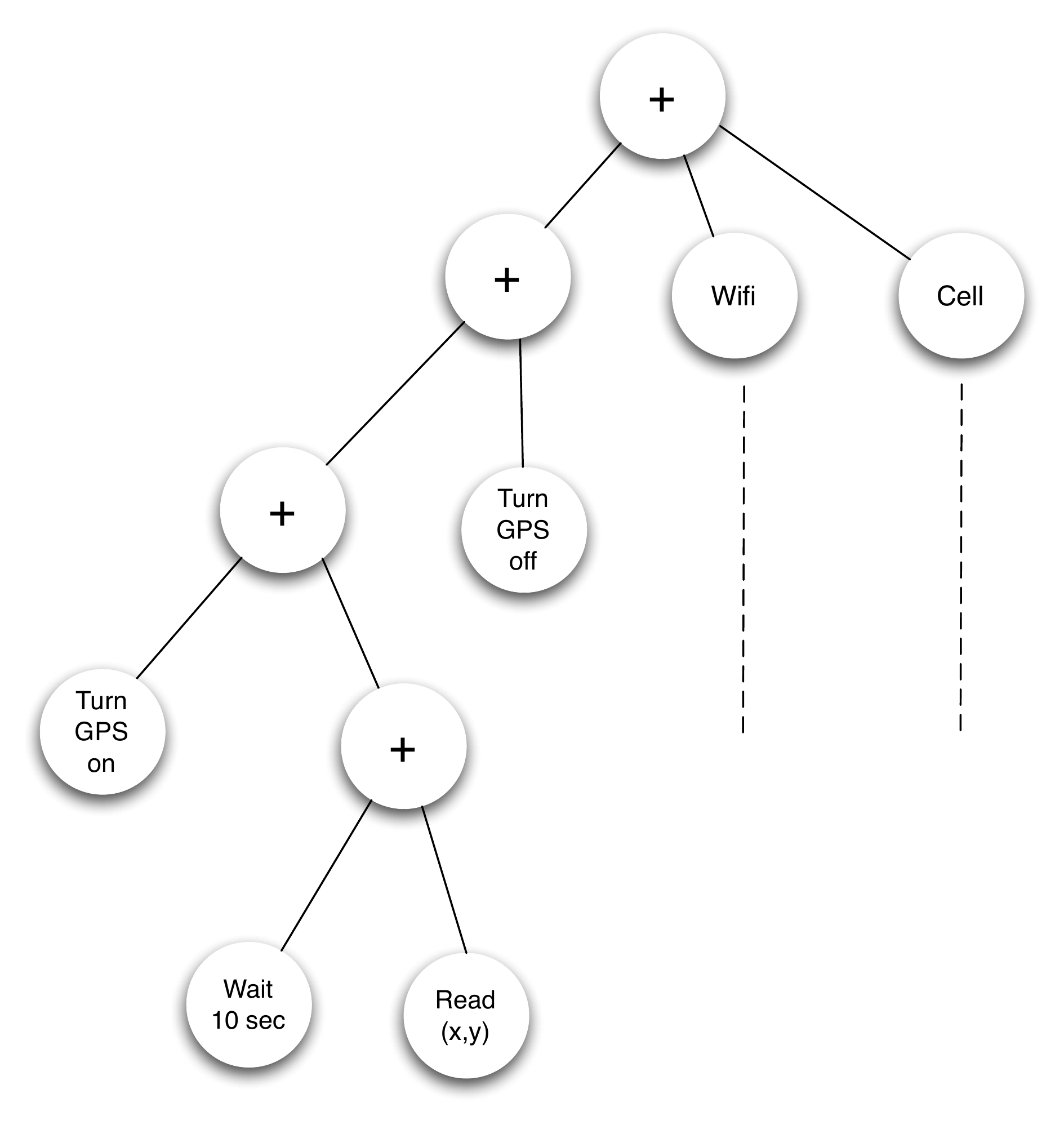}
\label{fig:TreeView}}
\subfigure[Combining functional logic blocks to reach peaks in the fitness landscape]
{\centering
\includegraphics[width=0.5\columnwidth]{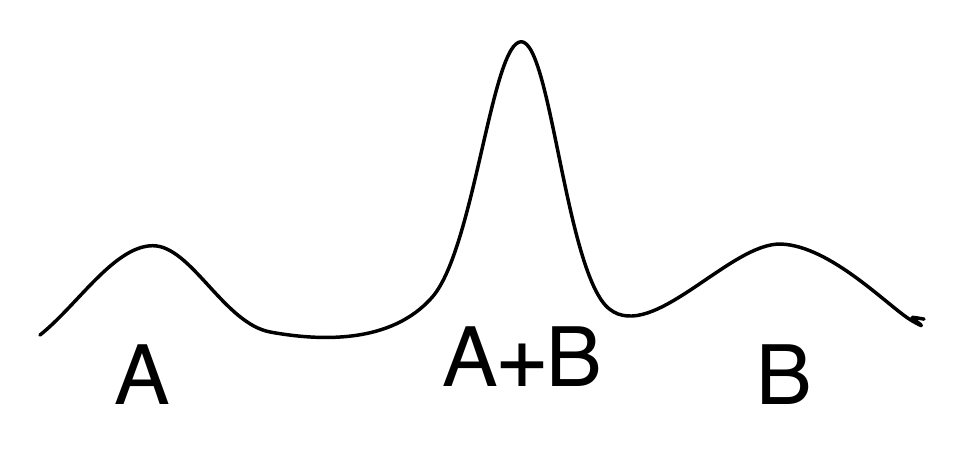}
\label{fig:combine logic}}
\caption{Online GP's logic representation supports sharing of functional logic blocks across islands}
\end{figure}
Neural networks typically learn the mapping between a set of input parameters and an output through an offline training process, and the resulting neural network is used for online predictions of new outputs according to new inputs. 
It is  possible to train an existing neural network topology in an online manner; however, such a fixed approach limits the capacity to share blocks of logic. It is also possible to evolve neural networks by representing their topology (and potentially interconnecting weights) as a genetic encoding. 
This permits the sharing of functional blocks represented by partial networks, yet it is much more likely that functional blocks will be arbitrarily split, meaning the output of a partial network may not match well as an input to another network. In contrast, online GP  ensures through syntactic checking that the output of a block of logic will always be acceptable to another block, which is arguably  beneficial in problems where solutions are naturally represented as programs (logical blocks). 

Consider the program tree representation in Figure~\ref{fig:TreeView} which provides a partial view of the energy-efficient localisation application in~\cite{cotillon}. The main goal of this application is to dynamically select among various positioning modalities on the smartphone to obtain accurate position estimate subject to energy limitations. The deeper branch in the tree represents learned GPS logic, while the deeper nodes of the other two shallow branches are omitted for simplicity. Let the logic in this GPS branch be functionality A, whose performance in the search space is indicated in Figure~\ref{fig:combine logic}. Assume there is another island that has evolved a separate program with functionality B. Because GP maps functionality to specific branches in the program tree, the sharing of genetic material between the two islands has a reasonable chance of combining the two functionalities A+B to reach much higher performance. A similar merging of such `blocks of logic' is significantly more challenging with other online learning approaches. 

Similarly, when sharing genetic material between subpopulations that have evolved in isolation, the syntactic checking within GP always ensures genetic material can be readily incorporated, unlike evolutionary neural networks where a high level of speciation could occur to the extent where any network from one subpopulation may not be compatible with any network in another subpopulation. Such a scenario would again be detrimental to collaborative evolution.


%

\subsection{Collaborative GP Personalisation}
Personalising complex applications through online learning may require long convergence time  and may have only partial contextual information. More complex personalisation may take a long time to converge as the search space grows, consuming a significant portion of the phone's processing cycles and energy.  Additionally, sensor data on a smartphones represents a single spatial sample of a physical property, which may not be representative of the broader context to which personalisation is required. For instance, if a mobile user requires personalised application response based on local weather, a smartphone placed in the sun can report ambient temperatures that are much higher than reality, biasing evolution toward non-representative context. 

The innate inter-communication capability between mobile devices lends itself to the Island Model implementation where each mobile device hosts a population and evolved programs can be serialised and shared (migrated) \citep{folino}. While intuitively one expects the parallel resources will expedite convergence, the Island Model is also known to generate better quality solutions \citep{Martin97}. 


We  argue that distributed logic evolution mitigates the shortcomings of stand-alone online personalisation by sharing genetic material among several smartphones and providing a broader context through increased spatial sampling.      Because smartphone users have a tendency to congregate together and possibly form small community-like groups,  co-located phones have a significant overlap in their context, which enables quicker and more effective personalisation by sharing logic among these phones. 

\section{Android Genetic Programming Framework}
\label{sec:agp}






Faster convergence towards desirable behaviour and learning shared distributed context are both strong motivators for collaborative smart phone personalisation. We choose to demonstrate this approach by extending the AGP framework to support the Island Model. In designing this collaborative learning method, we consider the following:

\begin{itemize}
\item Broadcast communication: we select broadcast best-effort communication interfaces for migrating programs to avoid maintaining connection information and to support graceful degradation. When a migrant does not reach its destination, evolution at the destination continues smoothly. Broadcast communication also provides scalability, where a single transmission of a migrant program can be received by multiple co-located phones. 
\item Flexible configuration: setting parameter values for the collaborative learning, such as the rate at which programs are shared (migration frequency), the number of migrating programs (migration intensity), and the type of migrating programs (mutated or crossover programs), depends on the application.  We aim to ensure flexibility in setting these parameters.
\item Compatibility: we aim to maintain well-defined application programming interfaces for AGP and to maintain full compatibility with the underlying Android OS.
\item Implicit context sharing: collaboration between phones is based on sharing logic rather than raw data, as the logic on a remote device will have evolved based on the perceived context at that device.
\end{itemize}

Prior to describing our collaborative smartphone personalisation approach on the basis of the above guidelines, we briefly revisit the Android Genetic Programming (AGP) framework. 

\subsection{Android Genetic Programming Framework}
\begin{figure*}[htp]
\centering
\includegraphics[width=0.8\textwidth]{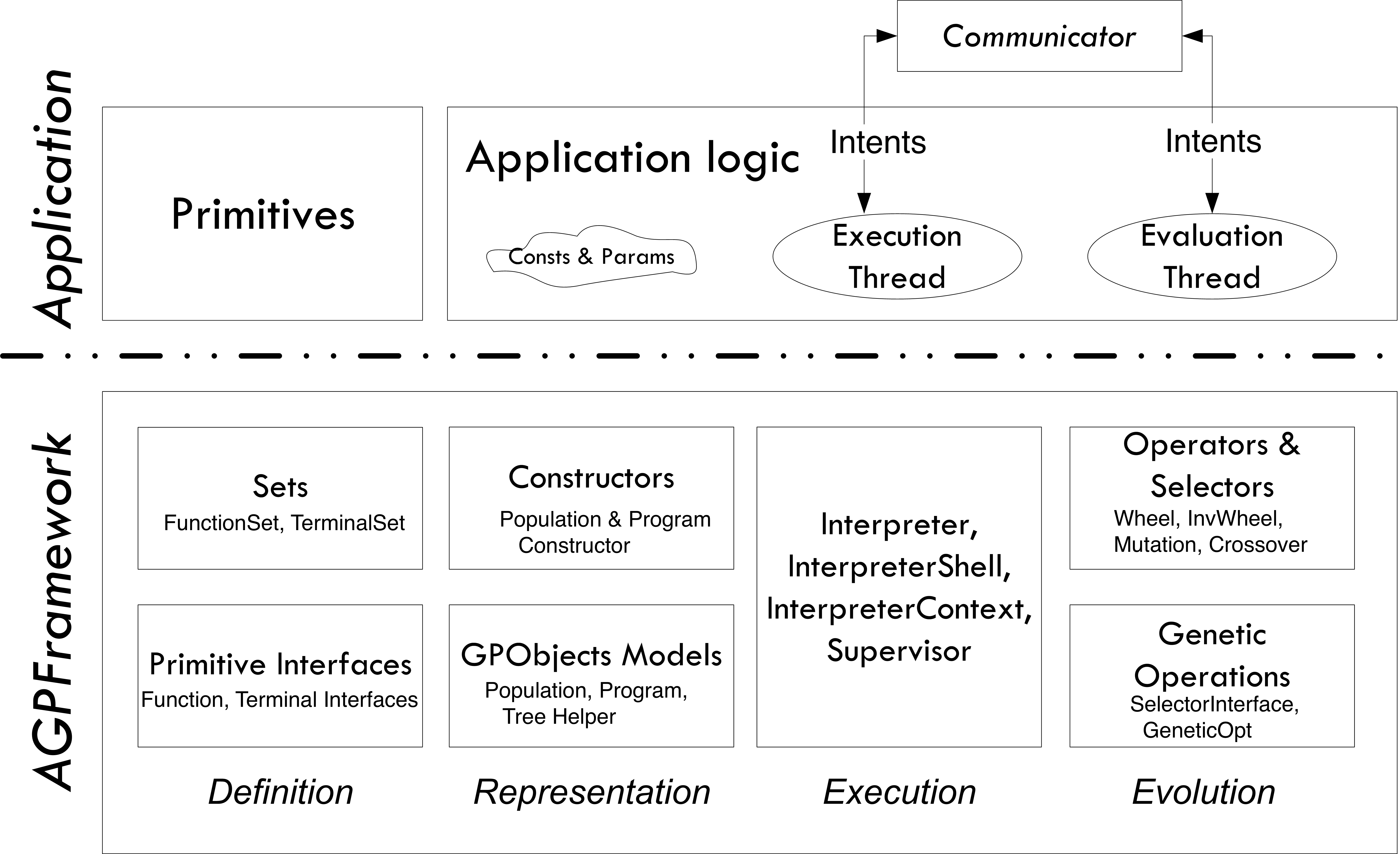}
\caption{Global Architecture of AGP}\label{fig:global}
\end{figure*}

The AGP framework \cite{cotillon} was developed as an easy-to-use genetic programming framework for evolving multi-objective, context-sensitive, adaptable applications on the Android phone operating system. Figure~\ref{fig:global} provides a high level view of the AGP framework. The core AGP framework provides the infrastructure for writing new GP applications for Android smartphones. 
For full details of the AGP framework the reader is referred  to ~\cite{cotillon}. The main purpose of AGP is to provide online GP infrastructure transparently to users for continuous smart phone personalisation. The original AGP framework supports only standalone GP evolution, which is why we extend it to support the Island Model as  Section~\ref{sec:island} describes. 

\subsection{Selectors and Genetic Operations}
When the evaluation thread is running alongside the execution thread, programs are evaluated and receive a fitness value. In GP, the programs that perform well are chosen to breed the next generation.  Selectors are organized following the Strategy Pattern in order to provide flexibility. This way, the developer can easily switch between standard selectors provided by AGP such as the Wheel selector, or create his own selector solution.

AGP then executes genetic operators on selected programs. In this regard, AGP currently supports two primary genetic operators widely used in GP, crossover and mutation, although it is extendible with more operators.

\subsection{Enhancements}
As GP is a stochastic process,   convergence towards desired performance can take several generations.    In order to ensure that convergence time does not strain energy and processing resources on smart phones, AGP includes two components, namely the helper and the supervisor, which enable the use of expert knowledge to apply constraints to AGP-generated programs.

\subsubsection{Helper}
\label{sec:helper}
The helper is called during the program generation process of any program builder. The developer can create one or several helpers for one application by implementing the AGP HelperInterface interface. Whenever a program is generated, AGP will refer to the helper evaluate() method to   check any correctness conditions that the program has to meet. For instance, in our geolocalization application (cf. Section \ref{sec:localization}), if a program doesn't call any location provider, we know that this program will be unable to locate the smartphone, so we can reasonably discard it without losing evaluation time, and therefore we save battery life. 

\subsubsection{Supervisor}
The supervisor runs during the program interpretation. It can check constraints on-the-fly, and kill the Interpreter Shell if the program goes out of bounds. For instance, the supervisor can control execution time. If the program execution is too long, the supervisor will automatically kill the program.

\subsection{Island Model}
\label{sec:island}
The Island Model is a parallelised genetic algorithm (GA) \cite{CantuPaz95} which employs multiple, distinct subpopulations (islands) in order to promote the genetic diversity of the complete system. 
Individual genomes can migrate from one subpopulation to another in an attempt to mix diverse genetic material to promote genetic diversity and alleviate premature convergence. 

\begin{figure}
	\centering
	\subfigure[Original AGP framework]
                {\centering
                \includegraphics[width=0.3\textwidth]{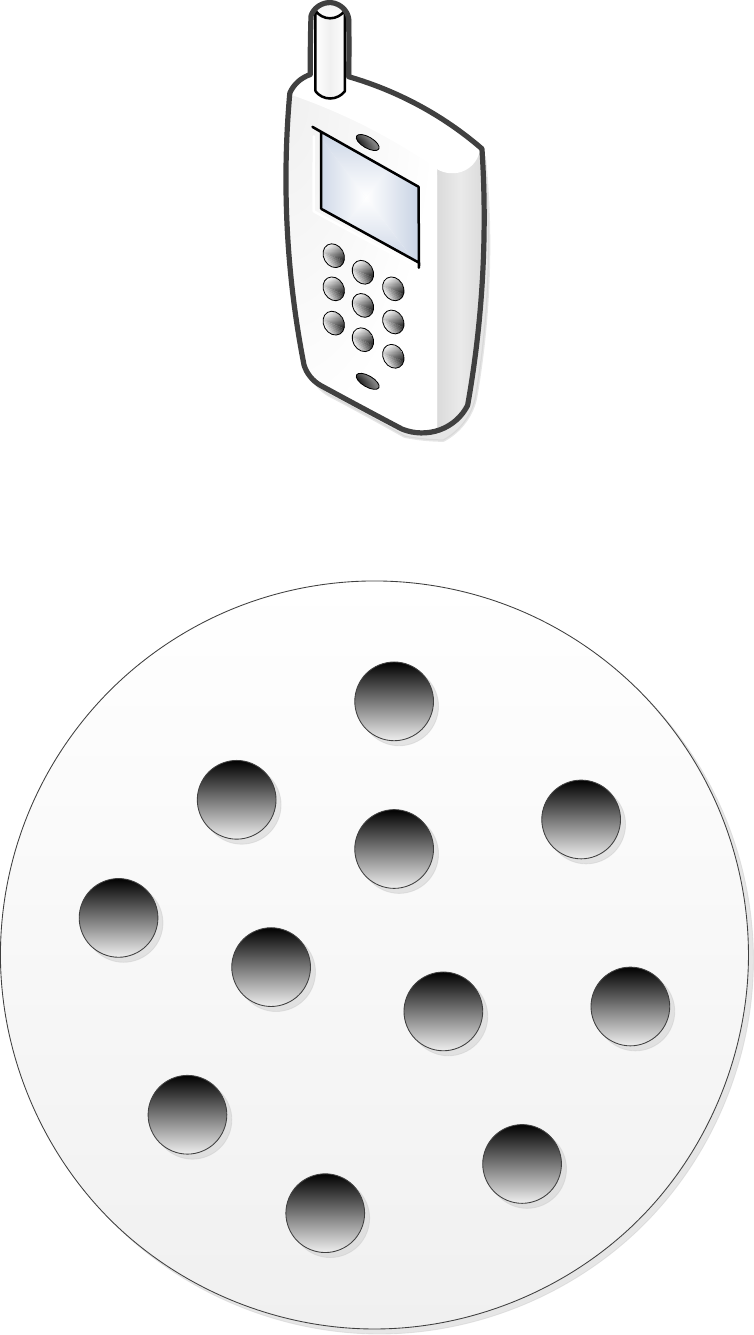}
                }
                \hspace{0.5cm}
       \subfigure[AGP framework with Island Model extension]
              {  \centering
                \includegraphics[width=0.3\textwidth]{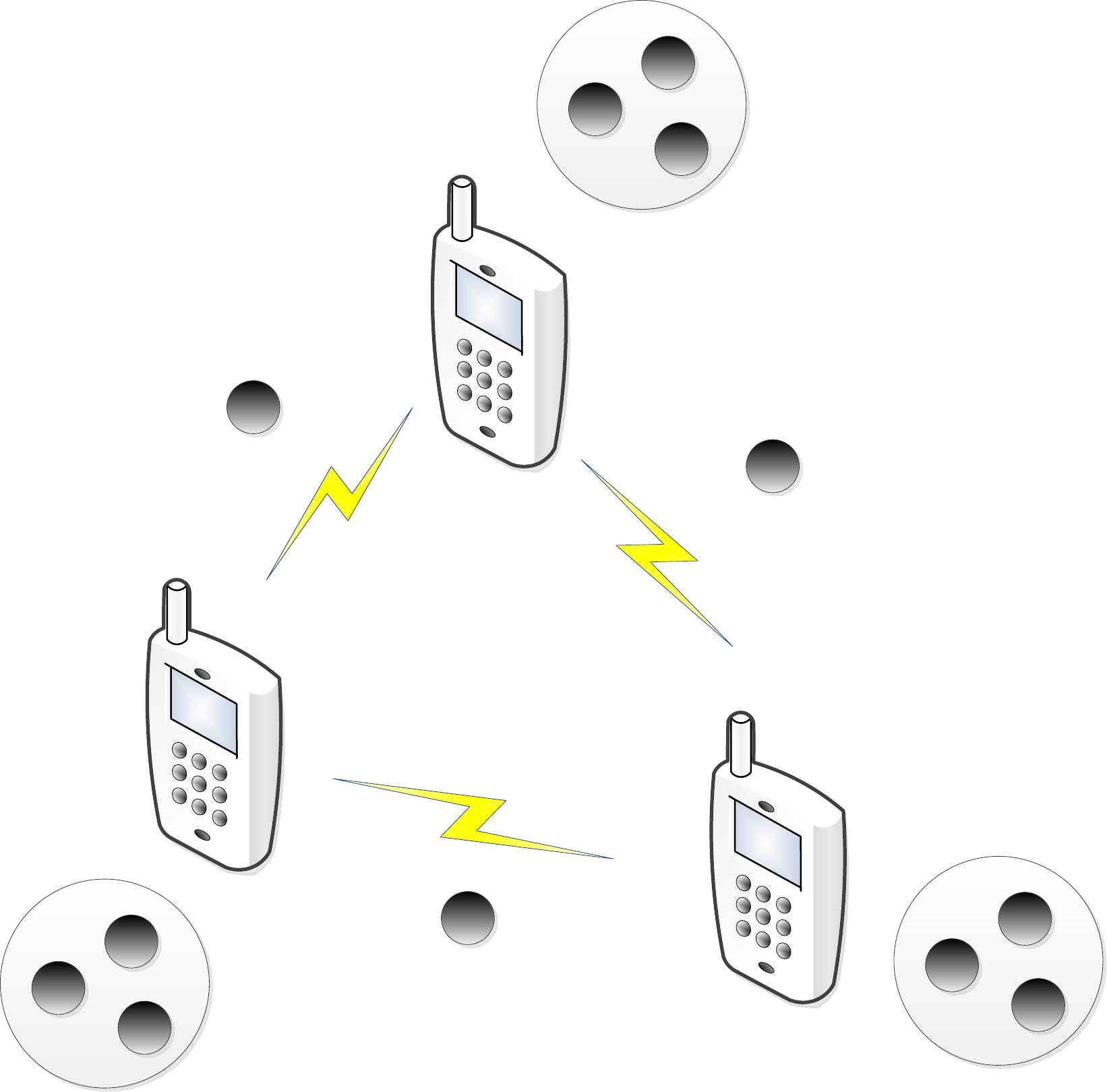}
                
     }
	\caption{Comparison of the original AGP framework and the Island Model extension. In the extension, each phone contains a local population that periodically exchanges genetic programs during migratory periods.}
	\label{AgpFrameworkIslandImplementation}
\end{figure}

The AGP framework creates a local population on each device, essentially an island with no migration inward (immigration) or outward (emmigration). However the inherent communications capabilities of mobile phones makes the Island Model an attractive extension to the AGP framework and also provides a mechanism for harnessing the collective resources of multiple devices for the benefit of all devices. 
These benefits can potentially manifest themselves as faster convergence times, more robust solutions and even more fit solutions. Figure~\ref{AgpFrameworkIslandImplementation} provides a synoptic comparison of the standalone AGP approach and the Island Model approach. 

We extend AGP to support the island model by allowing the island running on each phone to send and receive migrant programs from and to its local population. Since smartphones incorporate short-range wireless communication interfaces, such as Bluetooth and Wifi, we use these interfaces for the transmission and reception of migrant programs. The full implementation details of AGP's features that support the island model are available online.




\section{Performance Evaluation}
This section demonstrates the capability of smart phone personalisation through online GP using 2 example applications. 
\label{sec:perf}

\subsection{Standalone Personalisation}
We demonstrate standalone personalisation through AGP with applications on two different generations of Android devices: the early HTC Magic running Android 1.6 ; and the more capable Nexus S, embedding a dual-core processor with the recent Android Gingerbread (version 2.3).

\subsubsection{Google Reader Personalisation}
\label{sec:google}

\begin{figure}[htp]
\centering
\includegraphics[width=0.3\textwidth]{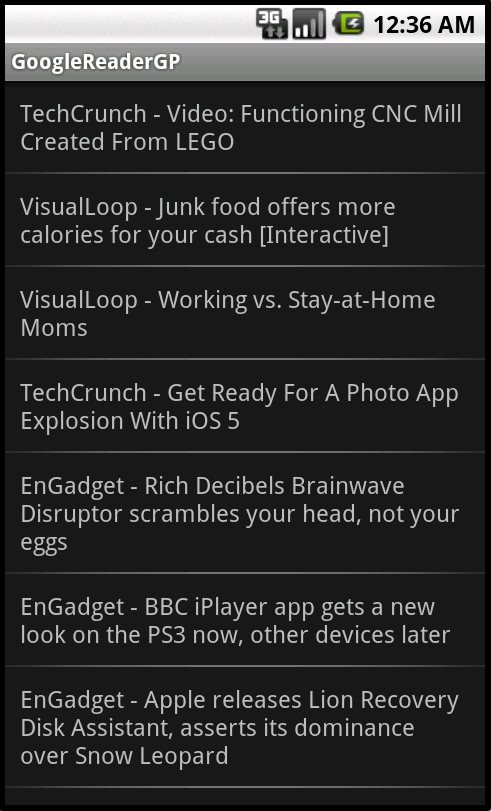}
\caption{Screenshot of the GRGP interface}\label{fig:GRGPInterface}
\end{figure}

Google Reader is a web-based aggregator released and maintained by Google. It works as a RSS feed reader, allowing users to get  the latest news from selected feeds. Many applications exist for  extracting news feeds from smartphones.  Preferences for reading news on a smartphone may depend on the  content type,  particularly its readability on a mobile device with a small screen. For instance, it is easy to read short text news whereas it is uncomfortable to look for long articles, comics, infographics or flash animations. Moreover, such content might quickly deplete the user monthly capped data plan.

Available Google Reader apps on the applications market reports news from all the feeds to which the user has subscribed. Even though some feeds are not easily readable on a mobile device, the application doesn't learn from the user reading habits and keeps proposing items from all the feeds.
Our Google Reader GP (GRGP) application takes advantage of the AGP framework to learn which feeds the user likes to read on their smartphone  within a given context. The feeds are taken from the user Google Reader account.

The application is kept simple for demonstration purposes: whenever the user wants to get news, she asks for a news report which executes a GP program and returns the latest and unread news from feeds selected by the program.

\paragraph{Fitness definition}
\label{sec:GRfitness}
The fitness definition is based on two sub-fitness functions:

\[Fitness = Fitness_{count}\times Fitness_{clicked}\]
\[
\label{eq:1}
Fitness_{count} =
 \begin{cases}
 \frac{Disp.\enspace news}{Desired\enspace qty.} & Disp.\enspace news \le Desired\enspace qty. \\
 1 & otherwise
 \end{cases}
\]
\[Fitness_{ clicked} = \left(\frac{Clicked\enspace news}{Displayed\enspace news} \right)\]


\normalsize
The news count refers to the minimum desired quantity of displayed news whenever the user requests a news report. This is a setting fixed by the user in the configuration menu.
The clicked news corresponds to the quantity of news read over the amount of news given in the report. In other terms, it evaluates the interest of the user in the displayed news.


\paragraph{Population evolution}
This experiment uses 5 program per population. Once every program in the population has  been executed, they all will have received a fitness value. It is  then time to generate a new population by using what  has been learned. We use elitism by copying the fittest program into the new population.  In GRGP, we use a typical roulette wheel selector. Thanks to selectors, evolution operators and tools embedded in AGP, we are able to clearly define our evolution strategy. Table \ref{tab:genOpt} gives basic commands to call  a basic strategy relying on mutations and cross-over done over the 3 best programs, which we refer to as highly ranked (HR) programs. We then perform mutations on the HR programs from the previous  generation to generated 2 mutated programs, which are included in the new population. The final 2 programs in the new population are crossovers of HR programs in the previous population. The maximum number of epochs in the simulation was set to 100. We set the maximum program depth to 3 to limit the program bloat, which is common in GP. As we have chosen problems with limited complexity to demonstrate the capabilities of AGP, this control mechanism was sufficient to avoid any bloat in our experiments.

\begin{table*}[t]
\caption{Basic commands for defining evolution strategy}\label{tab:genOpt}
\begin{center}
\begin{tabular}{ p{5cm} p{7cm}}
\hline\noalign{\smallskip}
\textbf{Action} & \textbf{Command}\\
\hline
\noalign{\smallskip}
Set a wheel selector with the 3 best programs & setSelector(WHEEL,``HR'', getnBest(3))\\
\hline
Set a wheel selector with the best program & setSelector(WHEEL,``Leader'', getnBest(1))\\
\hline
Generate 2 mutated programs & generate("HR", MUTATION, 2)\\ \hline
Generate 2 programs from crossover & generate("HR", CROSSOVER, 2)\\ \hline
Copy the best program into the new population & generate("Leader", COPY, 1)\\ \hline
Get the new population & getOutputPopulation()\\
\hline
\end{tabular}
\end{center}
\end{table*}



\begin{figure}[htp]
\centering
\includegraphics[width=1.0\textwidth]{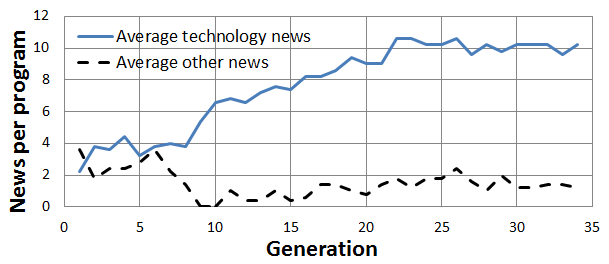}
\caption{The evolution of the number of news stories displayed in the standalone Google Reader application. }\label{fig:ndGRGPResults}
\end{figure}

\paragraph{Results}
We conduct our experiment with 7 sources from a real Google Reader subscription: 4 technology news websites (TechCrunch, TechLand, Engadget and Digital Trends), VisualLoop which gives fresh infographics, Break Videos for funny videos, and Business Green for latest green products. Even though the concerned user was interested in all those sources, he is used to read the technology news on his smartphone rather than the other sources providing longer articles or heavy media files,  as they are not optimised for  reading on smartphone.  We set the target number of news stories to 10. 

Figure \ref{fig:ndGRGPResults} reports the average of displayed news per program over the first 30 generations, where we observe convergence. In order to provide a  more readable graph, we grouped the news feed in two sub-groups: the 4 technology news, and the others (VisualLoop, BreakVideos, and Business Green). After 6 generations, our Google Reader using AGP learned the user preference for the technology news. However, it doesn't eliminate completely the diversity and keeps proposing some news from the other feeds yet with a lower likelihood.

As expected, GRGP  learns to provide the desired minimum quantity of news per report. Figure \ref{fig:ndGRGPResults} confirms this convergence by looking at the sum of technology and other news.

\begin{figure}[htp]
\centering
\includegraphics[width=1.0\textwidth]{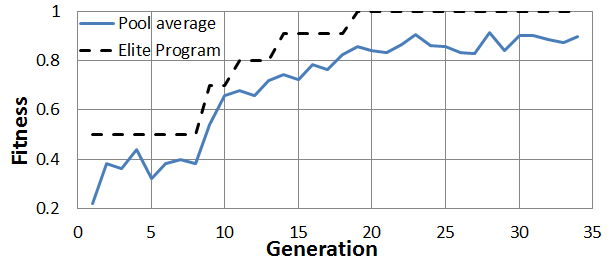}
\caption{Program fitness evolution for the standalone Google Reader application. }\label{fig:pfeGRGPResults}
\end{figure}

The program fitness evolution indicates that our evolution strategy does its job, and leads to an increase of the elite program fitness. In Figure \ref{fig:pfeGRGPResults}, the pool average represents the mean pool fitness (i.e. mean fitness of the 5 programs). Whenever the elite program gains in fitness, it subsequently leads to an increase of the pool average fitness.

\subsubsection{Context-aware localization}
\label{sec:gps}
\label{sec:localization}
 We now turn our attention to  adapting the localisation strategy running on the smart phone based on its context.   Localization is a complex problem as  there are multiple location services on a typical smart phone: GPS, which is  energy-hungry yet accurate and only works outdoors; WiFi, which  provides mid-level accuracy and energy consumption, and works when there are WiFi access points nearby; and   cell-based localisation that relies on cellular phone towers, which is energy-efficient yet inaccurate. The various localization options on smartphones requires consideration of their availability and their energy/accuracy trade-off~\citep{jurdak,jurdak13}.

As this problem depends on too many contextual constraints such as position, signal quality, and device energy profile, it is  difficult to design a context free algorithm  for selecting the best location service, particularly since the most accurate service depends on the user's current location and whether they are indoors or outdoors.  We  introduce a  context-aware application using AGP aimed to address this problem of selecting the best location service provider.   The focus here is on AGP's ability to solve multi-objective problems online.

\paragraph{Fitness function definition}
The fitness function used in our localization application reflects the common trade-off between energy and accuracy in the localization field \citep{lin}. We introduce two fitness metrics: the accuracy fitness and the energy fitness, which respectively quantify the accuracy and energy efficiency of the provided solution. As positions are dynamic, we evaluate fitness every second during the evaluation period (\emph{n} seconds). By the end of the evaluation time, we use the average of these subfitnesses to give a fitness to the program. The overall fitness is obtained by multiplying these subfitnesses. We choose this option as it discards any solution which doesn't provide any accuracy or could deplete the battery, while maintaining simplicity for the demonstration purpose of this paper.

\begin{eqnarray}
\scriptsize
Fitness = \frac{\sum_{i=1}^n Fitness_{Acc}(i)\times Fitness_{En}(i)}{n}
\end{eqnarray}

\paragraph{Accuracy fitness}
Most smart phones include several location providers, such as GPS, Cellular Network and Wi-Fi, that can be used alone or in combination.  When the application is learning, the evaluation process keeps all the location providers on,  which makes learning costly from an energy perspective. The application automatically picks the provider giving the most favorable accuracy. We call the output from this provider the best available position.
\begin{figure}[htp]
\centering
{\includegraphics[width=1\textwidth]{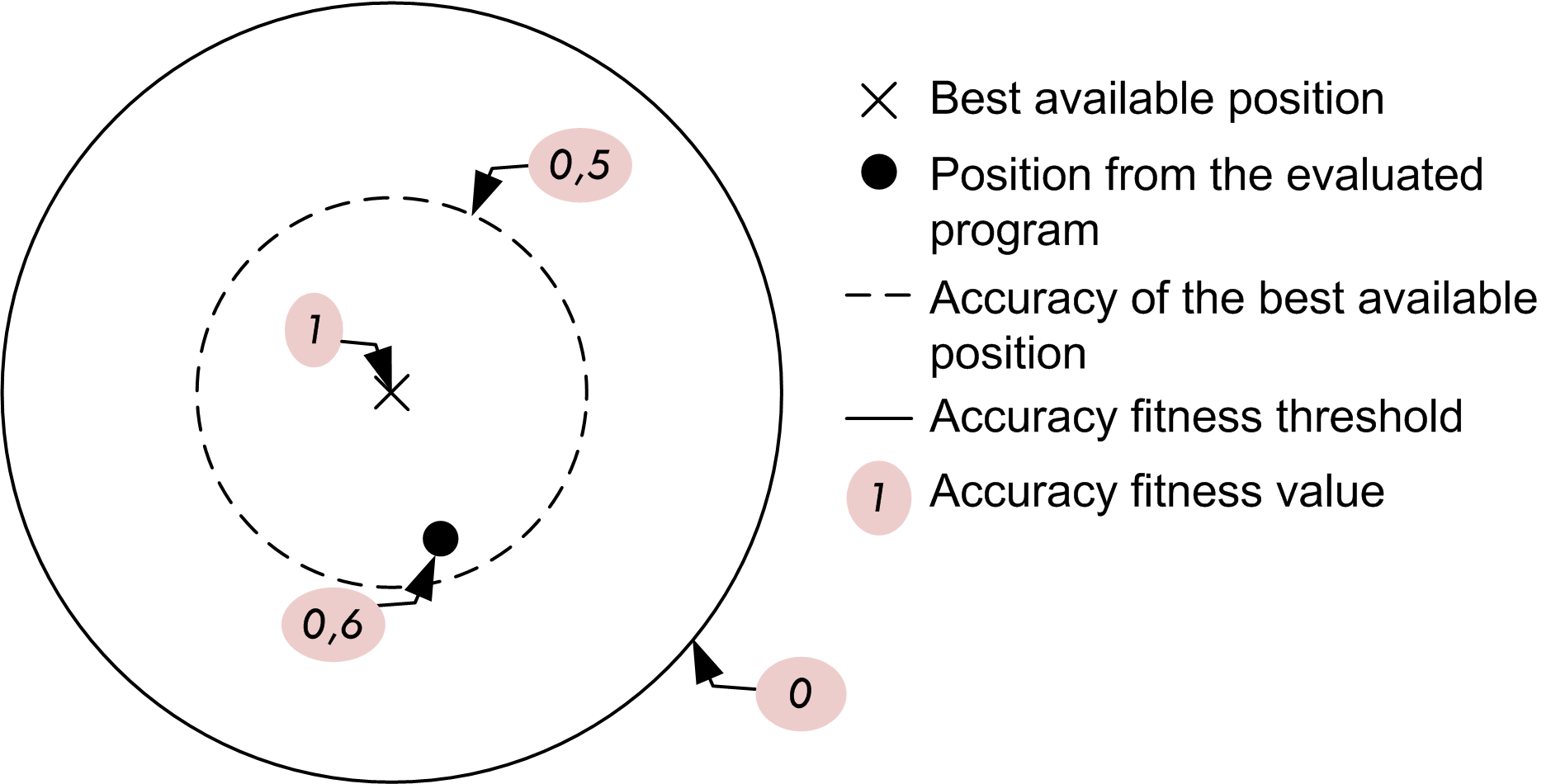}
\caption{Fitness accuracy definition for context aware localisation}\label{fig:accuracyFitness}
}
\end{figure}

We use this best available position as a reference to attribute an accuracy fitness to the position provided by the evaluated program (cf. Figure \ref{fig:accuracyFitness}):\\
- if program position is within the accuracy of the best available position, we attribute an accuracy fitness following a linear rule from 1 to 0.5\\
- if program position is outside the former, but within a circle of twice the accuracy of the best available position (accuracy fitness threshold), we attribute an accuracy fitness following a linear rule from 0.5 to 0.

\paragraph{Energy fitness}
We define energy fitness according to a basic rule: we want to provide  the best possible localization accuracy without depleting the battery by the end of the day as we assume users can charge their phones at the end of each day. We consider an average 1400 mAh battery capacity for the paper. To achieve our goal, the average power consumption should not exceed 63 mA (= 1400 / 22 hours). We define a day as 22 hours because we consider the phone as plugged for 2 hours per day.


We use the Android PowerProfile class to estimate the power consumption per chip. As an internal OS function used to compute the battery usage, this class isn't accessible through the Android Developer API. However, it can be readily obtained through the Java reflection mechanism. 

This enables AGP to assess the power cost of the evaluated program depending on the CPU usage and the chip used to locate the smartphone. For simplicity, we limit the energy fitness to a linear function, ranging from 1 for a idealistic case where the program doesn't cost any power to 0 for a program which requires more power than the one day energy budget.

\begin{figure}

\includegraphics[width=1\textwidth]{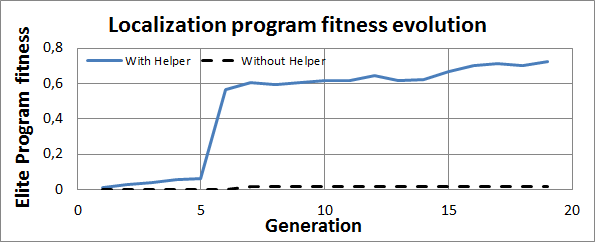}
\caption{Fitness evolution for context-aware localisation}\label{fig:localizationFitness}

\end{figure}

\paragraph{Results}
We conduct the experiment with populations of 12 programs, of which 1 is the fittest (elite) program from the previous generation, 4 are mutations, 5 are crossovers, and 2 are randomly generated programs. Mutations and crossover program are generated in the exact same way as for the Google Reader application, while the random programs are introduced to increase genetic diversity of the population. Because each program in this experiment requires much longer than the Google Reader Programs (for instance to acquire fixes from GPS), we set the experiment to terminate after 40 epochs. The program evaluation is limited to one minute to provide sufficient time for the GPS module to obtain a fix.
Our function set contains general operators such as addition, multiplication, and other application-specific operators: functions to switch location providers such as GPS, Wi-Fi or Cellular Network. 
Figure \ref{fig:localizationFitness} shows the framework ability to get a program localizing the smartphone. After a rough first solution, it converges to smarter programs able to provide more efficient and accurate solutions.

We also conduct an experiment to evaluate the benefits from the use of Helper.
Populations generated with the Helper provide a working solution in the first generation, and quickly have satisfying programs. On the other hand, populations generated without the Helper are stuck with non-working programs (zero fitness) for several generations. Then, they only get a slow evolution. It is mainly due to many programs which make no sense: they don't switch on a location provider or don't call any latitude nor longitude update.

\section{Collaborative Personalisation Experiments}
\label{sec:island_exp} Having demonstrated online smartphone personalisation through GP, we now evaluate collaborative personalisation on the same example applications. 
\subsection{Setup}
Both example applications use two Samsung Google Nexus S smartphones running Android 2.3 Gingerbread.  The population size for the collaborative personalisation experiments has  10 programs. The entire local subpopulation is replaced each generation whereby the fittest program (elite) is retained between generations. The three HR programs are then mutated to create three new programs, and the remaining programs are generated using crossovers applied to the entire breeding pool. We use  proportionate selection with respect to a program's fitness value, and we select emigrants at random without removing them from the local subpopulation. Immigrant programs are appended to the local subpopulation. Experiments are conducted using two different migration intervals (epoch length) \(i\) of 5 and 10 generations and three different migration rates \(r\) of 0.1, 0.2 and 0.3 of local population size --- i.e. 10\%, 20\% and 30\% of the local subpopulation are exchanged each migration period. We collect data about the genetic programs' fitness, size and tree depth for 20 generations and average it over 15 iterations to ensure accuracy and minimise the effect of the initial starting conditions.

The experiments use a migration schema, where genetic programs are randomly generated and added to the local subpopulation in lieu of the traditional migration between different subpopulations and smartphones. This enables the evaluation of the performance benefit versus energy consumed for the communication of genetic programs between phones. The genetic programs are randomly generated by an instance of the ProgramBuilder class and added to the subpopulation at a migration interval of \(i = 5\) and a migration rate of \(r = 0.3\).

Furthermore, for the Google Reader test application, we conduct further experiments using two different fitness landscape configurations: (1) a homogeneous fitness function over the two populations; and (2) a heterogeneous fitness function over the two populations. The differences between the fitness functions for the Google Reader application are simply the news sources that the user was interested in --- i.e. clicked. For example, in one population TechCrunch and Engadget news items are ``clicked'' and in the second population Break Videos and Digital Trends news items are ``clicked'' instead. Such differences in the fitness function are common in real world scenarios where each phone has a different inputs/constraints --- e.g. different user preferences, different context or different phone type. Thus our results using the heterogeneous fitness function provide a more realistic  evaluation of the Island Model's performance.

\subsection{Results}

\begin{figure}
	\centering
	\subfigure[Max Fitness]  {              \centering
                \includegraphics[width=0.6\textwidth]{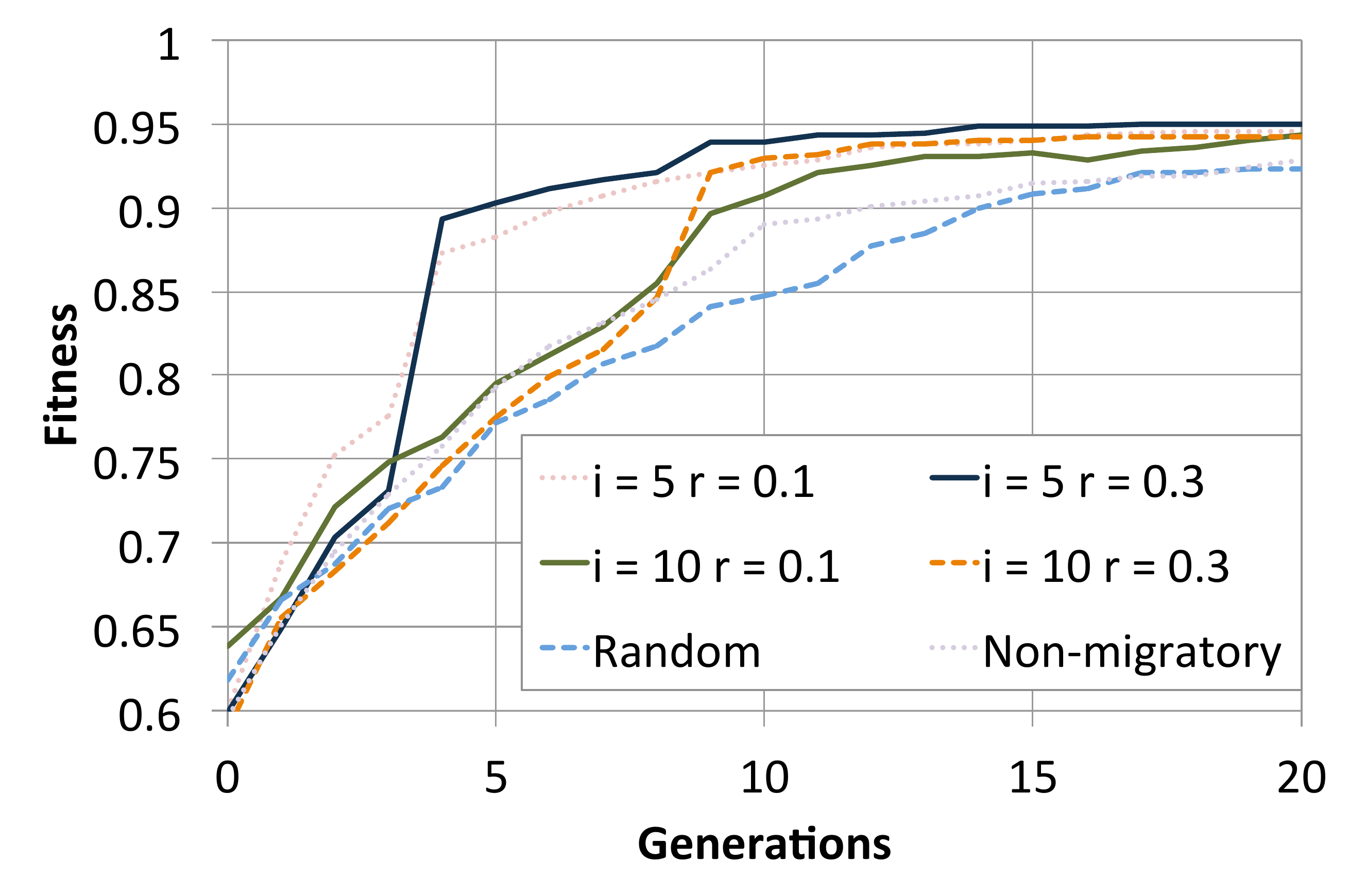}
                
                \label{fig:heteromax}}
        ~ 
       \subfigure[Mean fitness]{
                \centering
                \includegraphics[width=0.6\textwidth]{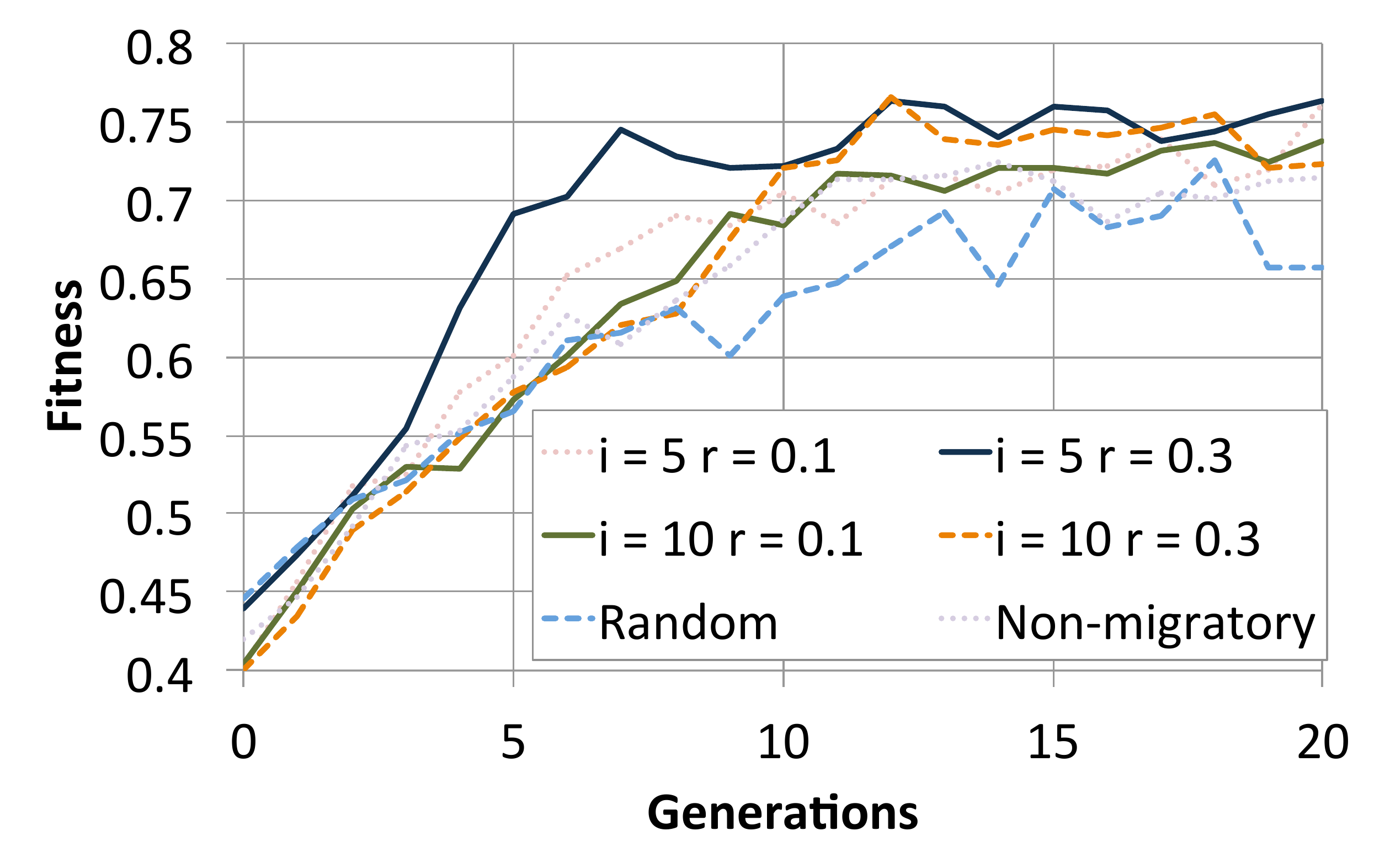}
          
		\label{fig:heteromean}
  }
	\caption{Fitness data for the Google Reader example application using a heterogeneous fitness function.}
	\label{fig:googlereader_hetero}
\end{figure}
\begin{figure}
	\centering
\subfigure[Max Fitness]  {
                \centering
                \includegraphics[width=0.6\textwidth]{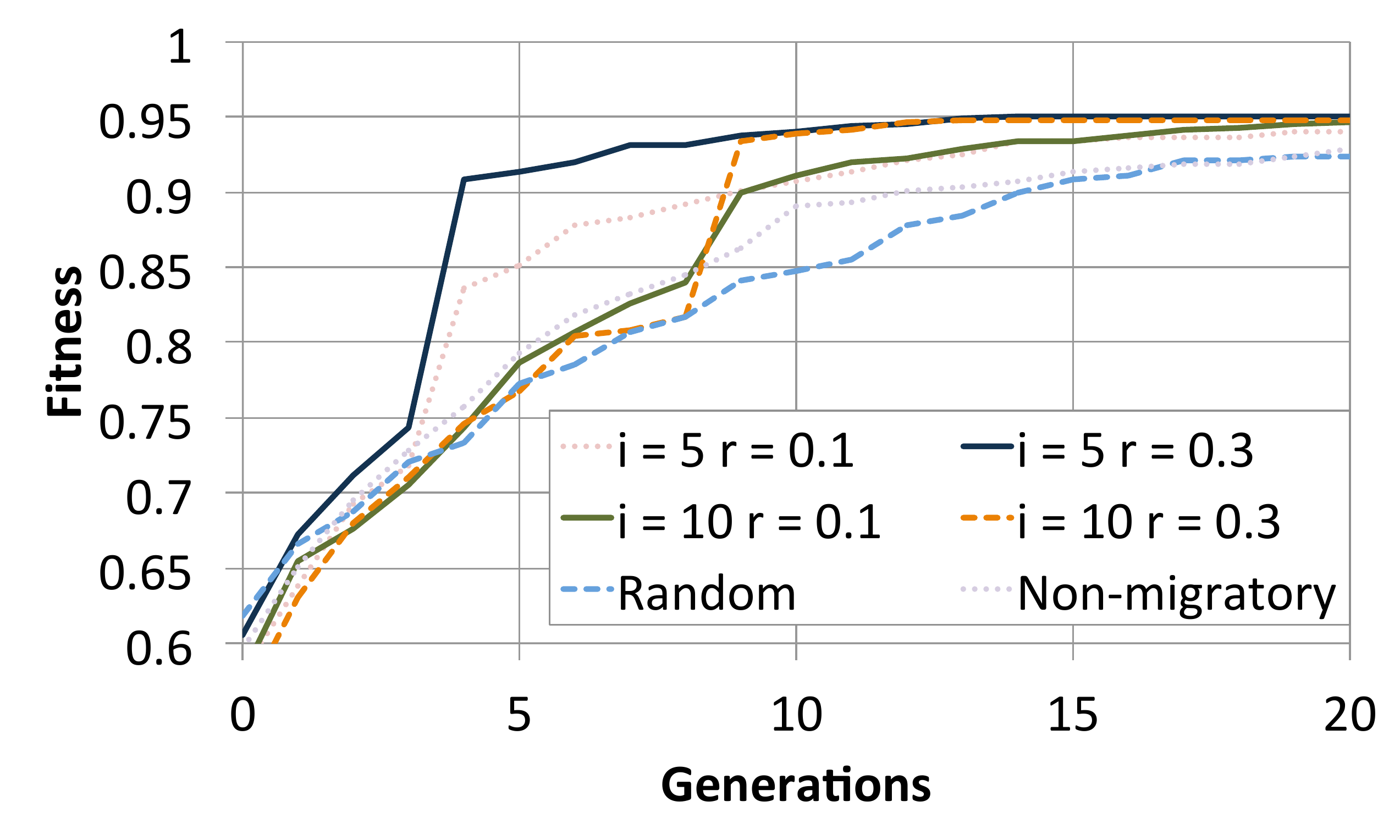}
        
                \label{fig:homomax}
   }
        ~ 
              \subfigure[Mean fitness]{
                \centering
                \includegraphics[width=0.6\textwidth]{Mean-homo-font}
               
		\label{fig:homomean}
       }
	\caption{Fitness data for the Google Reader example application using a homogeneous fitness function.}
	\label{fig:googlereader_homo}
\end{figure}
\begin{figure}
	\centering
	    \subfigure[Max fitness]{
                \centering
                \includegraphics[width=0.65\textwidth]{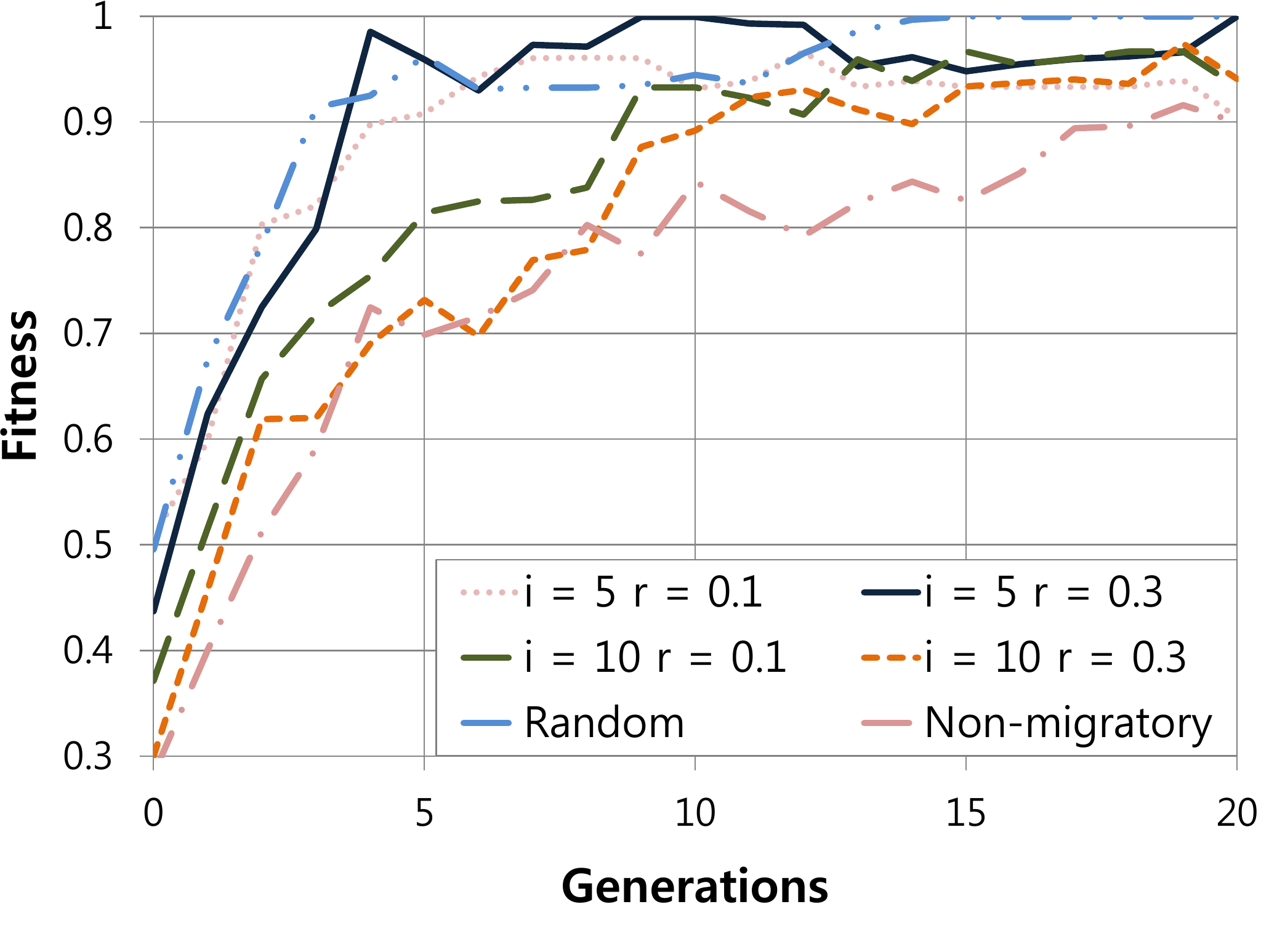}
               
                \label{fig:locmax}
      }
        ~ 
          \subfigure[Mean fitness]{
                \centering
                \includegraphics[width=0.65\textwidth]{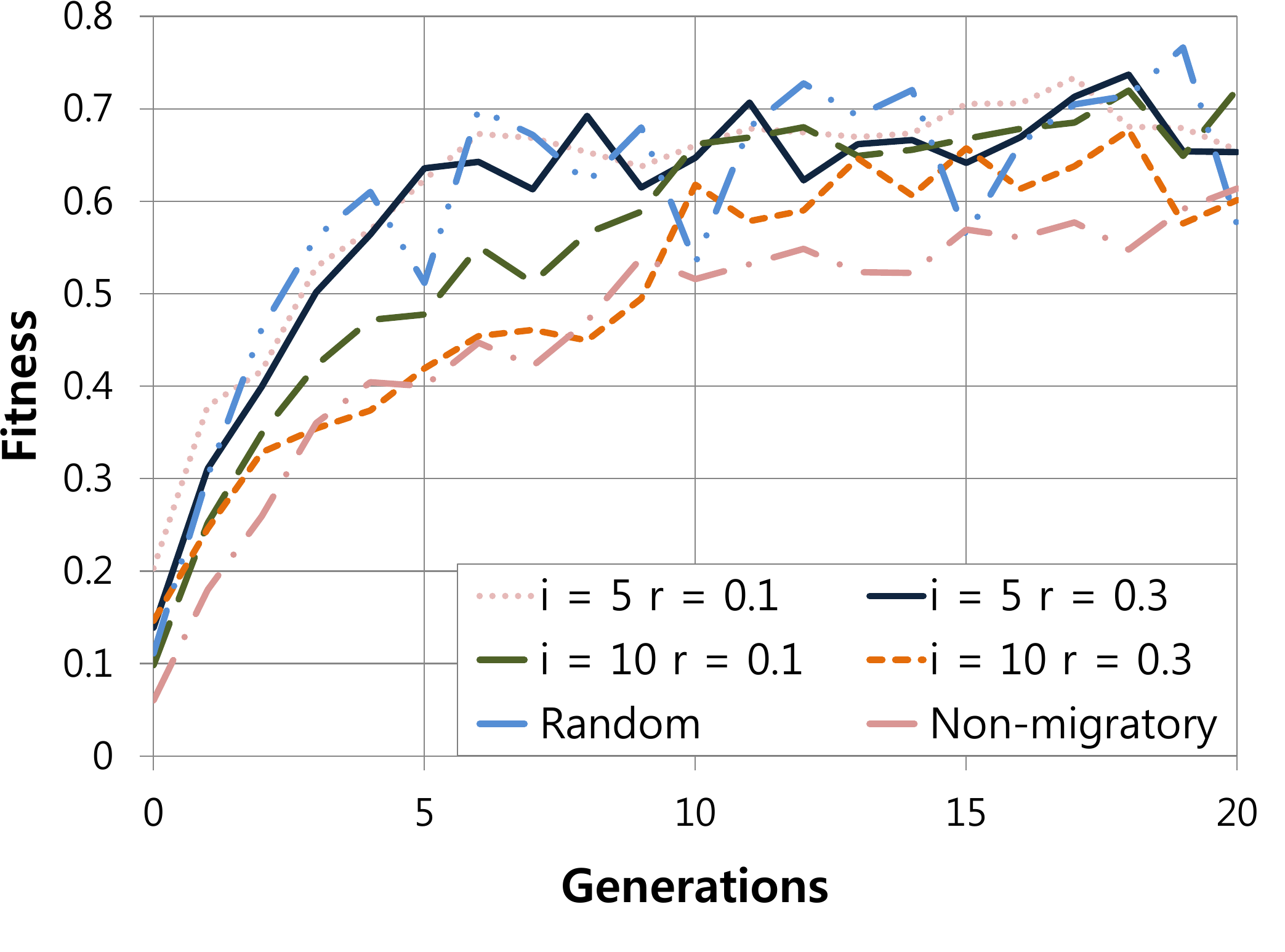}
    
		\label{fig:locmean}
}
	\caption{Fitness data for the localisation example application.}
	\label{fig:localization}
\end{figure}

Figure~\ref{fig:googlereader_hetero} shows the results from the experiments for the Google Reader application using a heterogeneous fitness function among the subpopulations and Figure~\ref{fig:localization} shows the results from the experiments conducted using the localisation example application. The figures show the experimental results for four of the six variations of the two migration parameters. In every tested configuration, the Island Model outperforms the non-migratory population model in terms of convergence time. For instance, in Figure~\ref{fig:homomax} shows that the elite program with migration reaches a fitness of 90\% by generation 4, while it takes until generation 12 to reach this fitness with the standalone model, representing a 66\% improvement. 

For the experiments using a homogeneous fitness function (c.f. Figure~\ref{fig:googlereader_homo}), we find that the fitness trajectory shows short exponential increases or spikes in the maximum fitness value in the generations immediately following a migration period, which can be seen at generation 4 for the experiments with i=5 and generation 9 for the experiments with i=10. This can be explained by the building block hypothesis \cite{Holland75}, where during the periods of isolated evolution the subpopulations evolve useful "building blocks". Subsequently when migration occurs between the two subpopulations, the good blocks  in the migrant programs are assimilated within the local subpopulation's genetic pool leading to the construction of larger and better novel building blocks and resulting in fitter genetic programs.

We can see similar results for the Google Reader application using the Island Model and a heterogeneous fitness function, which display the same characteristic short exponential increases in the maximum fitness value after each migration period. However, the introduction of the migrants typically  causes a temporary decrease in the average fitness followed by a rapid increase in the maximum fitness, supporting the results obtained by Martin et al \cite{Martin97}, M\"{u}hlenbein \cite{Muhlenbein91} and Belding \cite{Belding95}. This can be attributed to the heterogeneous fitness function or fitness landscape, influencing each subpopulation to explore a different area within the genetic search space, helping to promote genetic diversity between subpopulations.

Whilst still somewhat noticeable in the Google Reader fitness data, the effect of the initial starting conditions on the results is more pronounced for the localisation application. This is highlighted by the different starting locations of the mean and maximum fitness values for the non-migratory model and each different Island Model configuration demonstrated by the higher standard deviation and standard error values for the fitness data. This effect could be exacerbated due smartphone's context relevant to the objective function not being completely static during the evolution and execution of the genetic programs --- e.g. GPS reception and network reception and connection speed varying slightly over time and altering the fitness of executed genetic programs. It is recommended in future studies to use a larger sample size to mitigate this effect.

The results show there is no fitness benefit from the introduction of the randomly generated programs for the experiments using the Google Reader application. This contrasts with the results for the localisation application that demonstrate that the random programs have a potentially positive impact on the GA's performance. However, the average maximum fitness for the Google Reader application was much higher than for the non-migratory population model suggesting that the effect of the initial starting conditions is somewhat responsible for these results. We can also observe that there are significant dips in the average mean fitness immediately following the introduction of the randomly created programs.  Followed by a slightly larger increase in the subpopulations mean fitness as the junk genetic material introduced during the migratory period is discarded  through evolutionary processes. This can sometimes also result in a small increase to the subpopulations max fitness. 

The inconsistent performance due to the introduction of random programs can be attributed to the effect of random sampling having a small, but finite, chance of introducing more fit genetic material than in the current population. However, the probability of this occurring is much less than the probability of randomly selected programs from other subpopulations having a good fitness and this probability will continue to diminish as the average subpopulation fitnesses increase. This causes some inconsistency in the performance improvement of the random programs over the standard non-migratory population model. Nonetheless, these results are intriguing and accord with the intuition that any attempts to maintain or increase genetic diversity will typically have a positive influence on the population's evolution. 

Based upon these results, an interesting idea for future work is the epigenetic selection of terminals, such as the strategy suggested in ~\cite{Salustowicz}, to govern the creation of genetic programs. Essentially, grammars defined by probabilistic trees to select the function and terminal sets that are ultimately instantiated, could be dynamically evolved. For example, if some function \(A\) has terminals \(B\) and \(C\) as its children and is shown to have a high probability of a high fitness, then this bias could be used when generating new programs, such that there is a higher chance for any instance of the function \(A\) to have the terminals \(B\) and \(C\) as its children. 

\section{Discussion and Conclusion}
\label{sec:conc}

The benefits of the Island Model over the standard non-migratory population model are well known \cite{Muhlenbein91,Belding95,Martin97,Turner98} and this is clearly demonstrated in the obtained results. There is a very obvious increase in learning speed that can be attributed to the effects of migration when employing the Island Model. This is observable in the evolutionary trajectories of the fitness scores in both the Google Reader and the localisation example applications. Notably, the mean (pool) fitness decreases immediately after a migratory period, due to the typically lower fitness of the immigrant programs, however on occasion, these dips are then followed by a short and sometimes exponential increase in the elite fitnesses of each subpopulation as newly diverse and useful genetic material is incorporated into local subpopulation.

The experiments conducted with the Google Reader example application using a heterogeneous fitness function between subpopulations displayed a similar performance benefit to those using a homogeneous fitness function. Martin et al \cite{Martin97} suggest that this is due the different fitness functions promoting genetic diversity between the different subpopulations, consequently improving the exploitation and exploration of the genetic search space. This demonstrates that the heterogeneous nature of the pervasive computing environment --- i.e. different users and preferences, different smartphone properties, helps rather than hinders the performance of the Island Model, allowing the Island Model to more efficiently leverage the available processing resources compared to the traditional non-migratory population model.

Finally, it must be noted that some privacy concerns have been raised about collaborative techniques, specifically about fine-grained localisation \cite{Wicker2012}. However, genetic programming is well placed to help preserve users' privacy by sharing logic containing variables and not needing to share the actual values of the variables (such as location, energy, etc). Furthermore, the logic shared may or may not represent the user's actual preference due to the stochastic selection of which programs are migrated and finally there is no need to reveal the source of the program, so ideally the programs would be broadcast anonymously. 

Overall, the results of this study show the significant potential of the combination of the Island Model with the AGP framework for achieving collaborative learning of dynamic user personalisation in pervasive computing contexts. While we have tested the Island Model on only 2 phones and on relatively simple applications as a proof of concept, we expect the benefits of parallel resources and improved population diversity as the number of participating islands (phones) increases to be able to address more complex applications and expedite convergence times \citep{Martin97,Niwa98}.

The symbolic nature of GP means that logic can be readily seeded and is an intuitive choice where control of device's resources is typically performed programatically. It should be noted, however, that adaptive behaviour may not be desirable for all interactions. For example, the user interface should have a consistent feel across applications and adhere to the device or OS recommended UI design recommendations. However, within this constraint, adaptive behaviour may provide a method to overcome consistently undesirable application (i.e. fixed logic) behaviour. The current configuration employs a fixed population structure which produces a number of likely low performing programs to be evaluated every generation even after the system has converged. This means that there will always be some residual unexpected or unpredictable behaviour, however this exploration aspect is a necessary evil in order to maintain adaptability to significant changes in user preferences and also to refine performance over time. 


User satisfaction is mainly driven by  AGP's capability to quickly adapt to context and provide suitable solutions. To achieve this goal, AGP needs a learning process which can have an efficient energy profile. As the end-user is also  sensitive  to early battery depletion, introducing a  mechanism to  tune  this trade-off between battery and learning would help maintain satisfaction.  An interesting direction for future work would be to implement a mechanism  to configure AGP  for intensive learning or a low energy consumption mode depending on context.

Another area for future investigation is to extend the context aware localisation application and AGP to include other sensor streams that could aid localisation. For  example, the developer could also introduce other providers such as a contact-logging beacon method \citep{jurdak,jurdak13}, or an accelerometer-assisted localisation algorithm~\cite{jurdak12}. Finally, an interesting direction for future work is to explore the use Pareto GP algorithms for multi-objective problems to evaluate its potential benefits over compounded fitness functions that combine several objectives.

In summary, this paper has proposed collaborative personalisation through online GP. We have demonstrated the capability for smart phone personalisation through online GP, and have quantified the benefits of collaborative personalisation with the Island Model. We believe this study  represents a key first step towards versatile online personalisation in the growing smart phone market.
\bibliographystyle{plainnat}
\section*{Acknowledgements}
The authors would like to thank Brano Kusy for his valuable inputs in realising this work. This project was supported by the Sensors and Sensor Network Transformational Capability Platform at CSIRO.

\end{document}